\documentclass[runningheads]{llncs}
\usepackage[ruled,vlined,linesnumbered]{algorithm2e} 
\usepackage{graphicx}
\usepackage{subfig}
\usepackage{comment}

\begin{document}
\title{Relevance feedback strategies\\ for recall-oriented neural information retrieval}

\author{Timo Kats\inst{1}\orcidID{0000-0003-1650-1814} \and
Peter van der Putten\inst{1}\orcidID{0000-0002-6507-6896} \and
Jan Scholtes\inst{2}\orcidID{0000-0003-0534-2491}}
\authorrunning{Timo Kats, Peter van der Putten, Jan Scholtes} 

\institute{LIACS, Leiden University, P.O. Box 9512, 2300 RA  Leiden, The Netherlands\\ \email{t.p.a.kats@liacs.leidenuniv.nl, p.w.h.van.der.putten@liacs.leidenuniv.nl}
\and
Maastricht University, Minderbroedersberg 4-6, 6211 LK Maastricht, The Netherlands\\ \email{jan.scholtes@maastrichtuniversity.nl}} 

\maketitle              

\begin{abstract}
In a number of information retrieval applications (e.g., patent search, literature review, due diligence, etc.), preventing false negatives is more important than preventing false positives. However, approaches designed to reduce review effort (like ``technology assisted review") can create false negatives, since they are often based on active learning systems that exclude documents automatically based on user feedback. Therefore, this research proposes a more recall-oriented approach to reducing review effort. More specifically, through iteratively re-ranking the relevance rankings based on user feedback, which is also referred to as relevance feedback. In our proposed method, the relevance rankings are produced by a BERT-based dense-vector search and the relevance feedback is based on cumulatively summing the queried and selected embeddings. Our results show that this method can reduce review effort between 17.85\% and 59.04\%, compared to a baseline approach (of no feedback), given a fixed recall target.
\keywords{high recall information retrieval \and dense-vector search \and relevance feedback}
\end{abstract}

\section{Introduction}
In a diverse set of information retrieval applications, reaching certain minimal recall thresholds is key, when identifying which documents in a large corpus are relevant to an information need. For example, in intelligence and law enforcement use cases, searching for adverse effects of medicines, due diligence, and legal search applications, preventing false negatives is often more important than preventing false positives \cite{Song}. But, manually reviewing all documents exhaustively is cost intensive and error-prone \cite{Piramuthu}. Hence, technology is often used to reduce review effort when retrieving information.

However, technologies aimed at reducing review effort (like ``technology assisted review") can create false negatives, since they often rely on active learning systems that exclude documents automatically based on user feedback \cite{Grossman}. 

Therefore, this research aims to evaluate a more recall-oriented approach for reducing review effort when cumulatively identifying relevant documents. For this, we combine two major components: textual similarity and relevance feedback. Here, the textual similarity is used to produce relevance rankings, whilst relevance feedback is used to iteratively improve those relevance rankings based on user feedback. Hence, our first research question is formulated as follows: what text similarity methods are most suitable for relevance feedback? Our second research question builds on that and reads: to what extent can relevance feedback help to reduce review effort?

For the textual similarity component, we evaluate TF-IDF and BERT-based dense-vector representations. For the relevance feedback component, we compare text-based and vector-based feedback strategies. Moreover, we experiment with different levels of textual granularity in both components. In the text similarity component this is done through paragraph-based document rankings and in the relevance feedback component this is done through feedback amplification. 

This combination of representation methods, feedback strategies and levels of granularity is technically and methodologically novel, and leads to substantial improvements in results. Given these two components, our proposed method reduces review effort between 17.85\% and 59.04\% when compared to our baseline approach of giving no relevance feedback, given a recall target of 80\%.

The remainder of this paper is structured as follows. Section~\ref{sec:background} discusses related work. Next, we describe our methodology (section~\ref{sec:methodology}) and experimental setup (section~\ref{sec:setup}). In section~\ref{sec:results} we share the results of our experiments, and the paper ends with a discussion section (section~\ref{sec:discussion}) and conclusion (section~\ref{sec:conclusion}).

\section{Background}
\label{sec:background}

In this section we discuss work related to the two main components of our method: text similarity methods and relevance feedback strategies. For the text similarity methods we  discuss term-based and context-based approaches. Next, for the relevance feedback strategies, we review text-based and vector-based strategies. 

\subsection{Term-based similarity methods}
Term-based similarity methods compute text similarity through taking the common features between two pieces of text into account \cite{Chandrasekaran}. As a result, these methods don't incorporate contextual language properties like homonymy or synonymy in their similarity computations. 

A simple and commonly used \cite{Hampp} implementation of term-based similarity is Jaccard similarity \cite{Jaccard}, which computes text similarity based on how many features two texts have in common divided by the total number of features across both texts. The implementation of this similarity method is based on set algebra (using the intersection and the union operators). The formula for this is given below. Here, $A$ and $B$ refer to the set of unique words for both documents.

\bigskip
\begin{equation}
    Jaccard(A,B) = \frac{|A \cap B|}{|A \cup B|} = \frac{|A \cap B|}{|A| + |B|}
\end{equation}
\bigskip

However, a shortcoming of Jaccard is that frequent terms (that are more likely to match) within a document are unlikely to be distinctive or important \cite{Chai}. To deal with this shortcoming, there are two approaches. The first approach is based on manually filtering out frequent words based on a predefined list of words that are known to be common in a given language. These words are often referred to as ``stop words". Removing stop words when searching for textually similar documents tends to have a beneficial effect \cite{Chai}. Hence, we filter out stop words in our experiment. 

The second approach is based on diminishing the value of words that are common in a corpus \cite{Chai}. An advantage of this method compared to filtering stop words is that it's more dynamic. Certain words that are common within a specific context (e.g., the word ``patient" in medical data) might not be included in predefined lists of stop words.

This approach is implemented in our TF-IDF-based similarity method through  \textit{inverse document frequency}. Here, terms are given a measure of uniqueness by dividing the number of documents in total by the number of documents that have a specific term. The formula for this metric is given below. Here, $D$ refers to the number of documents in the dataset whereas $d$ refers to the number of documents that contain term $t$. As a result, terms that appear in many documents (and are therefore less unique) are given a diminished value \cite{Sanchez}.

\bigskip
\begin{equation}
idf( t, D ) = log \frac{ |  D | }{ 1 + | \{ d \in D : t \in d \} | }
\end{equation}
\bigskip

\subsection{Context-based similarity methods} 
In this research we also use context-based similarity methods. In contrast to the term-based similarity methods mentioned previously, these methods do incorporate a form of semantic meaning. This is based on the ``distributional hypothesis" \cite{Gorman}, which is built on the idea that ``words that occur in the same contexts tend to have similar meanings" \cite{Chandrasekaran}. Given this hypothesis, this research uses pre-trained word embeddings that provide vector-based representations of texts. This enables us to compute the textual similarity based on the similarity between the vectors (e.g., with cosine similarity) \cite{Chandrasekaran}. 

In this category of pre-trained word embeddings, we are specifically focused on BERT 
\cite{Devlin}, 
\cite{Vaswani}.  
The transformer-based architecture of BERT allows it to capture context-sensitive word properties (like homonymy and synonymy) in its embeddings. 
There are versions of BERT made for specific tasks. For example, Sentence-BERT (SBERT) is a BERT model specifically made for measuring text similarity using the cosine distance metric \cite{Reimers}. The key advantage SBERT has over the other BERT models for text similarity tasks is its reduction in computational overhead. The researchers found that for finding the most similar pairs in a collection of 10.000 sentences, BERT would take $\approx$65 hours whereas SBERT would take 5 seconds. Moreover, it enables a similarity search between larger bodies of text (like sentences and paragraphs) through mean-pooling the word embeddings.

\begin{figure}[tpb]
    \centering
    \includegraphics[width=7cm]{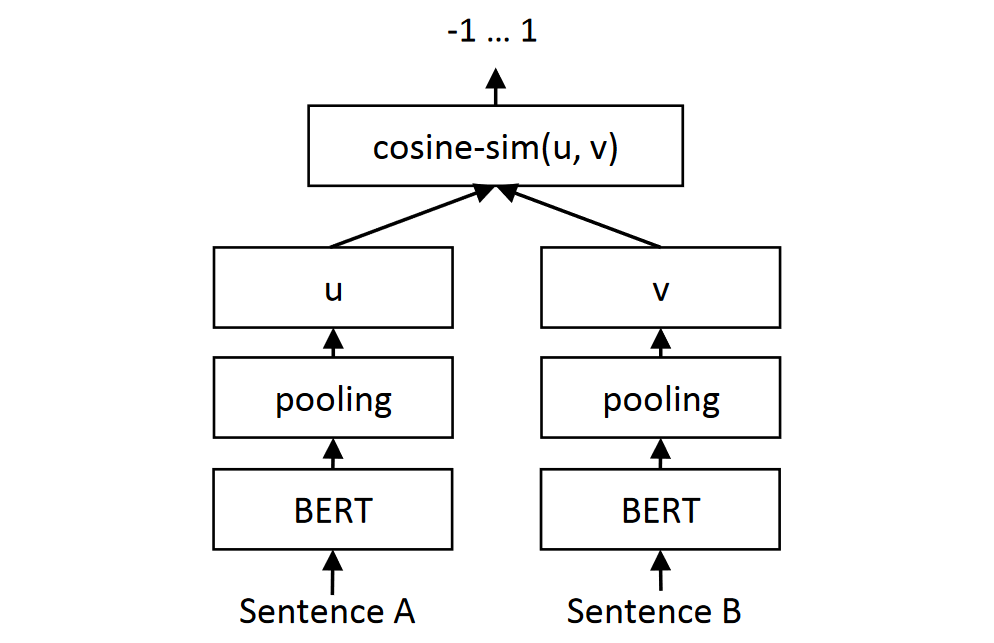}
    \caption{Simplified architecture of SBERT \cite{Reimers}}
    \label{fig:sbert-architecture}
\end{figure} 

A simplified overview of this architecture can be found in Figure \ref{fig:sbert-architecture}. Here, the (word-level) BERT embeddings are first mean-pooled to create paragraph or sentence level embeddings. Thereafter, these embeddings are compared through their cosine similarity. Note, for the mean-pooling process, text that exceeds 384 words in length is truncated \cite{Reimers}.  As a result, this approach can't be implemented on a level of text granularity that exceeds this amount. 

\subsection{Text-based feedback strategies}
Relevance feedback refers to changing the relevance ranking through user feedback, generally in multiple iterations \cite{Manning}. A commonly used text-based relevance feedback strategy is ``keyword expansion". Here, keywords from the selected search results are appended to the query. A frequently used implementation of this approach is based on the \textit{inverse document frequency} mentioned earlier in this section. Here, the underlying assumption is that more unique words are more valuable to the search. In this research we will use this strategy for selecting keywords to expand our queries with. 

\subsection{Vector-based feedback strategies}
Assuming vector representations of the query and search results are available, user feedback can also be applied to the queried vector directly. This is referred to as vector-based pseudo-relevance feedback. In general, there are two commonly used methods for this \cite{Li}. First, the queried vector can be averaged with the positive search results. Second, the queried vector can be summed with the selected search results. Both strategies are implemented in our research. 

A commonly used variation of averaging query vectors is based on Rocchio's method for relevance feedback \cite{Li}. The high-level idea of this method is to move the query vector towards the selected vectors through assigning different weights to selected and queried vectors \cite{Rocchio}. The version of Rocchio implemented by most researchers today \cite{Cai} \cite{Arampatzis} differs slightly from the original method, since it omits the negative feedback (i.e. non-selected documents) from the formula. As a result, this version of Rocchio can be seen as a weighted average between the (original) queried embedding and the (averaged) selected embeddings. 

The weight of the queried embedding ($\alpha$) and the weight of the averaged selected embeddings ($\beta$) can be set by a user.  Still, the default/consensus values most research adheres to is $\alpha=0.5$ and $\beta=0.5$ \cite{Arampatzis}. Hence, we use Rocchio with those parameter values as a baseline method in our experiment.

\section{Methodology}
\label{sec:methodology}

\begin{figure}[tbp]
    \centering
    \includegraphics[width=\textwidth]{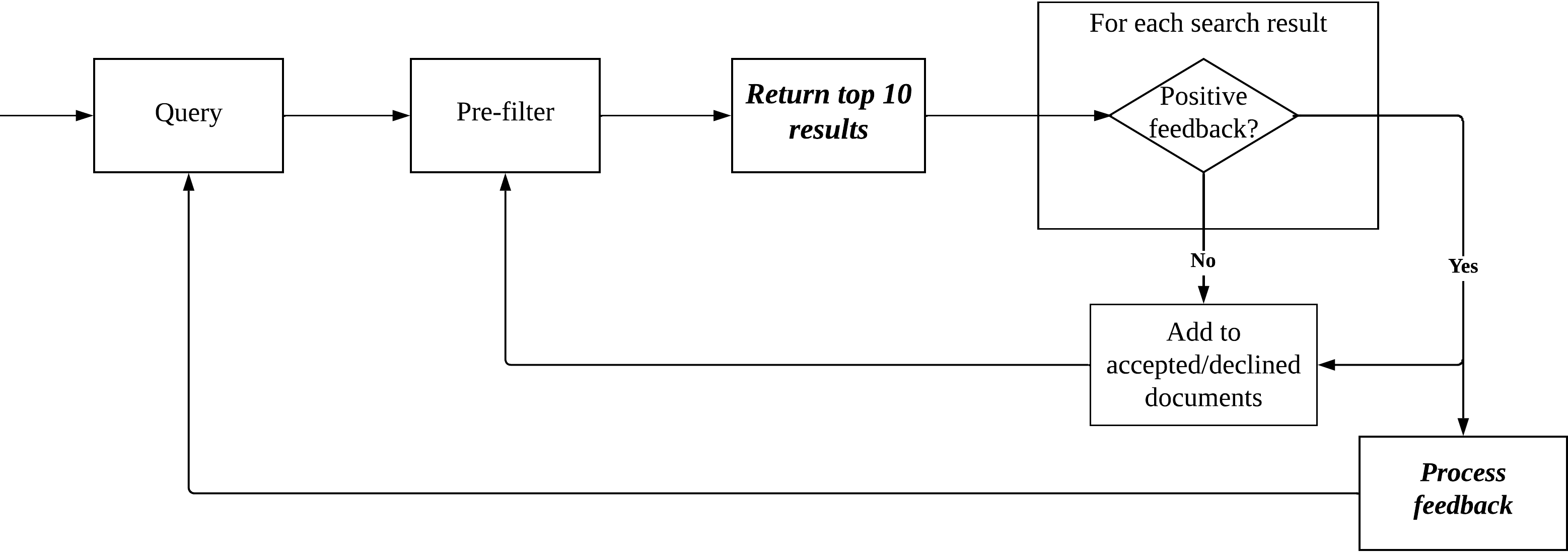}
    \caption{Flowchart of the method}
    \label{fig:overview}
\end{figure}

In this research we evaluate different relevance feedback strategies and text similarity methods with the objective of reducing review effort. As shown in Figure \ref{fig:overview}, the method for accomplishing this is based on iteratively presenting the user with a set of (10) results to accept or decline. Thereafter, the accepted documents are used to improve the query (and consequently results) for the next iteration.

For returning the results (i.e relevance ranking) we experiment with different text similarity methods, levels of textual granularity, and ranking methods. These methods are explained in the first and second subsections. Next, for processing the feedback given after each iteration, we experiment with different feedback strategies. These are explained in the third and fourth subsections.

\subsection{TF-IDF-based text similarity}
Our baseline method is based on TF-IDF, which vectorizes text based on multiplying the frequency of a word in a document with the \textit{inverse} frequency of that word in the entire data set. Hence, TF-IDF looks at the ``uniqueness" of a word instead of just the frequency. The result of TF-IDF is a sparse vector where all the words that exist in a document are given their TF-IDF value (words that don't appear in a specific document have zero values).

In our research we use TF-IDF to find similar documents using ``MoreLikeThis" (MLT). In summary, MLT queries the terms from a document (that have the highest TF-IDF values) individually using the document index. The documents returned by these queries are ranked based on combining their MLT scores, which is defined as the sum of the TF-IDF values of the matching terms between the queried document and the returned document \cite{Lucene}.

Note, we are aware that a commonly used implementation of TF-IDF to identify textually similar texts is based on computing the cosine distance between the TF-IDF vectors. However, computing the cosine distance between \textit{n} vectors  results in quadratic time and space complexity ($O(n^2)$). Hence, we use MLT instead.

\subsection{BERT-based text similarity}
Our third similarity method is based on Sentence-BERT (SBERT) embeddings. These embeddings can be used to find similar texts using ``dense-vector search" (DVS). The distance metric used for this is cosine similarity, which computes the similarity between two vectors (i.e. embeddings) based on the cosine value of the angle between the two vectors \cite{Rahutomo}. This angle is computed through dividing the dot-product (which is the sum of products from both vectors) by the length of the vectors. This means that the potential values of this similarity measure range from -1 (completely opposite) to 1 (completely similar). The formula for this is given below. 

\bigskip
\begin{equation} 
    {cosine\ similarity}= \cos (\theta)= {{\bf A} {\bf B} \over \|{\bf A}\| \|{\bf B}\|} = \frac{ \sum_{i=1}^{n}{{\bf A}_i{\bf B}_i} }{ \sqrt{\sum_{i=1}^{n}{({\bf A}_i)^2}} \sqrt{\sum_{i=1}^{n}{({\bf B}_i)^2}} }
\end{equation}
\bigskip

In order to find embeddings with a high cosine similarity (or low cosine distance) in a large dataset efficiently we use Hierarchical Navigable Small Worlds (HNSW) based vector search \cite{Malkov}. HNSW is an algorithm that finds the \textit{k} most similar documents to a query with logarithmic time and space complexity (O(log(N)). 

\subsection{Paragraph-based document rankings}
Because relevant information can be exclusive to a specific part of a document \cite{Adebayo}, we conduct experiments on two levels of text granularity: document and paragraph. Here, the paragraph level means that we query and retrieve paragraphs instead of documents. However, for the relevance ranking we only consider documents. As a result, the experiments on the paragraph level require a document ranking to be derived from the returned paragraphs. For this, we define two different paragraph-based document rankings.

The first ranking method is based on taking the highest ranked paragraph of a document in the ranking as the overall document ranking. For example, say we return 6 paragraphs from 3 unique documents in the following order: $\{d_1, d_2, d_2, d_3, d_3, d_3\}$ where $d_i$ refers to a paragraph from document $i$. Then, our document ranking will be as follows: $\{1, 2, 3\}$. Note how the ranking from the first paragraph of a document determines its position in the document ranking. 

The second ranking method is based on counting the number of paragraphs per document in the ranking, and using that to rank the documents. For example, say we return the same 6 paragraphs in the following order: $\{d_1, d_2, d_2, d_3, d_3, d_3\}$ where $d_i$ refers to a paragraph from document $i$. Then, our document ranking will be as follows: $\{3, 2, 1\}$. Note how in contrast to the previous ranking method, the number of paragraphs per document determines the ranking. 

\subsection{Baseline feedback strategies}
In this research we have three baseline methods for the relevance feedback experiments. The first baseline method is to re-use the original queried embedding. If a user accepts/declines documents, the ranking of documents remains constant. This is referred to as ``no feedback" or ``original". 

The second baseline method is based on keyword expansion. In keyword expansion, each iteration a number of keywords from the documents with \textit{positive} pseudo-relevance feedback is selected and appended to the original query using an OR operator. As a result, the selected keywords serve as a pre-filter for the original query where the documents should contain at least 1 of the collected keywords to be considered. In our research, this selection is done based on the (highest) \textit{inverse document frequency} value. 

\subsection{Vector-based feedback strategies}
Since the queried and collected texts have vectors (e.g., BERT or TF-IDF), the relevance feedback methods are based on vector operations. These vector operations are based on summing and averaging the vectors. For this, we experiment with both cumulative (i.e. include the queried vector in the average/sum) and non-cumulative feedback (i.e. exclude the queried vector in the average/sum). 

Also, for text similarity methods implemented on the paragraph level, relevance feedback can be amplified to the document level. If a given paragraph receives positive relevance feedback, then that feedback can be extended to other paragraphs that have the same parent document. In our research this will be referred to as ``amplified feedback" (or ``amp" in tabular formats). Note, feedback amplification is not applicable to any of our baseline methods. 

\section{Experimental setup}
\label{sec:setup}

This section provides an overview of the experiments and our data (and preprocessing). The first experiment focuses on comparing the performances of different text similarity methods. The second experiment focuses on implementing the best performing similarity method using different relevance feedback strategies.

\subsection{Data and preprocessing}
Our research uses the RCV-1 v2 \cite{RCV-2} dataset in all the experiments. This dataset was made public in 2005 by Reuters News and consists of 806784 news articles. Due to hardware constraints, we did not use the complete dataset in our experiment. Instead, we randomly sampled 300 articles per topic. For this, we sampled 15 (unrelated) topics that are equally split in train, validation, and test. Moreover, besides these topics we also sampled 4 topics that share the same parent topics. This set will be referred to as the ``ambiguous" set. 

As for preprocessing, in TF-IDF we filter out stop words, numbers, and convert the text to lowercase. The stop word list used for this is publicly available on GitHub\footnote{https://github.com/stopwords-iso/stopwords-en}. For the SBERT-based experiments, we only remove numbers and special characters.

Finally, for the experiments on the paragraph level, we first split the text into sentences using the \verb|<p>...</p>| tags in the XML files from RCV-1 v2 dataset \cite{RCV-2}. Next, a paragraph is created through concatenating every 3 adjacent sentences of a document (and the remainder). For SBERT the number of words in a paragraph can't exceed 384. Hence, we verified that the paragraphs in the topic sets do not exceed that limit.

\begin{table}[t]
\centering
\begin{tabular}{|l|l|l|l|}
\hline
\textbf{Model name} & \textbf{Size} & \textbf{\#Dimensions} & \textbf{Speed (sentences/sec)} \\ \hline \hline
all-mpnet-base-v2   & 420 MB        & 768                 & 2800                                    \\ \hline
all-MiniLM-L12-v2   & 120 MB        & 384                 & 7500                                    \\ \hline
all-MiniLM-L6-v2    & 80 MB         & 384                 & 14200                                   \\ \hline
\end{tabular}
\caption{Selected BERT models' characteristics according to \cite{Reimers}}
\label{tab:bert-models}
\end{table}

\subsection{Text similarity methods}
For the text similarity methods, we compare two approaches: MLT (which is based on TF-IDF) and DVS (which is based on SBERT). For both methods, we iterate through our set using ``query by document" (QBD). Each time we query a document/paragraph to return other documents/paragraphs that belong to the same topic. 

For MLT, we set two parameters. First, the minimum document frequency for words to be considered (minDf). Second, the maximum document frequency for words to be considered (maxDf). Due to limited computing resources, we didn't conduct a full grid search to find the optimal values for these parameters. Instead, we conducted a manual search on the train and validation sets and used the average of the most performant parameter values on our test (and ambiguous) set. Note, we conduct these experiments on the paragraph \textit{and} document level.

For DVS, we don't have any parameters to set. Hence these experiments are conducted directly on our test set and our ambiguous set. However, we do experiment with three different pre-trained SBERT models. These are selected based on being the general-purpose models in the SBERT documentation \cite{Reimers}. An overview of these models can be found in Table \ref{tab:bert-models}. Note, due to SBERT's maximum context length, we conduct these experiments on the paragraph level only.

Finally, we evaluate the similarity methods based on (the macro averaged) recall, precision and F1 scores. For these metrics, the definition of a ``true positive" is a returned document that is of the same set as the queried document. This positive set is based on the annotations of the dataset. For example, if we query a document annotated as ``sports", then the document returned (as similar) should also be annotated as ``sports" to be considered a true positive. 

\subsection{Relevance feedback}
For our relevance feedback experiments we implement a form of pseudo-relevance feedback. Here, the feedback is based on the same definition of a true positive as mentioned earlier. 

An overview of the layout of the experiment can be found in Algorithm \ref{alg:feedback}. Note how each iteration the already collected/declined documents are filtered from the search. Also, note how the query is updated each iteration based on the pseudo-relevance feedback. In our implementation of this experiment, we collect 10 documents/paragraphs each iteration. 

Finally, for performance evaluation, we record the average number of iterations needed to achieve exactly 80\% recall (which is a commonly used threshold in information retrieval \cite{Hedin}).

\begin{algorithm}[tb] 
\SetKwInOut{Input}{input}\SetKwInOut{Output}{output}
\SetKwComment{Comment}{// }{}
\Input{paragraph, maxRecall}
\Output{Iterations needed to achieve recall}
\DontPrintSemicolon
\caption{Pseudo relevance feedback experiment}
\label{alg:feedback}
\upshape{iterations} $= 0$ \\
\While{\upshape{recall(acceptedDocuments)} $\leq$ maxRecall}
{
    \upshape{filter} $=$ acceptedDocuments + declinedDocuments \\
    \upshape{results = query(paragraph, filter)}   \tcp*{Returns top 10 results.} 
    \For{\upshape{result \textbf{in} results}}
    {
        \If{\upshape{feedback for result is positive}}
        {
            \upshape{paragraph = processFeedback(result)} \\
            \upshape{acceptedDocuments += result}
        }
        \Else
        {
            \upshape{declinedDocuments += result}   
        }
    }
    \upshape{iterations} $+= 1$
}
\Return iterations
\end{algorithm}


\section{Results}
\label{sec:results}

This section is an overview of the results from our experiments. In the first subsection, we discuss the results of the individual text similarity methods. In the second subsection, we discuss the results of the different relevance feedback strategies.

\subsection{Text similarity methods} 
For our TF-IDF-based approach, the manual search on the train and validation sets resulted in the parameter values of maxDf=0.8 and minDf=0 on both the paragraph and document level. Hence, the TF-IDF related results are based on these parameter values. Next, for the DVS experiments, \textit{all-mpnet-base v2} was the best performing pre-trained model. Hence, the results of this model are shown in this section.

\begin{figure}[p]
    \centering
    \includegraphics[width=\textwidth]{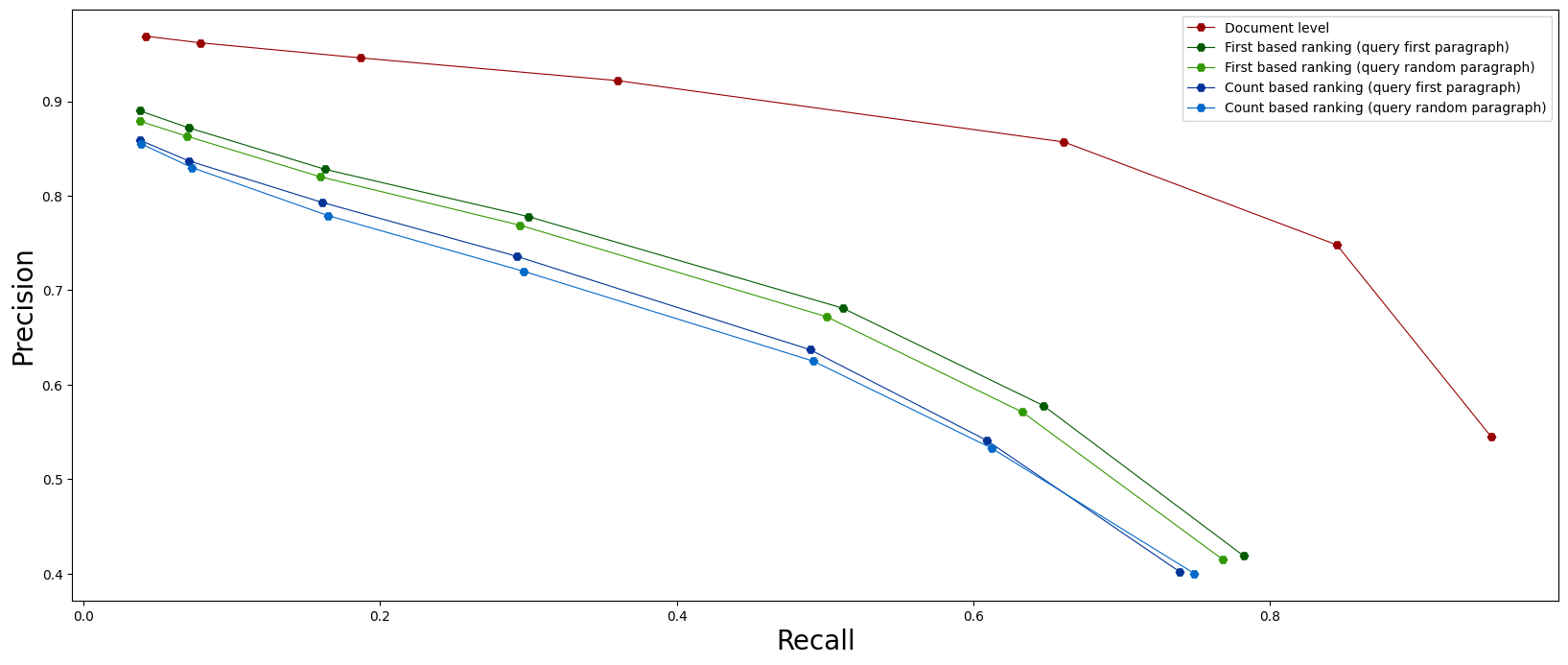}
    \caption{Different configurations for the TF-IDF-based approach on the test set}
    \label{fig:tfidf}
\end{figure}

\begin{figure}[p]
    \centering
    \includegraphics[width=\textwidth]{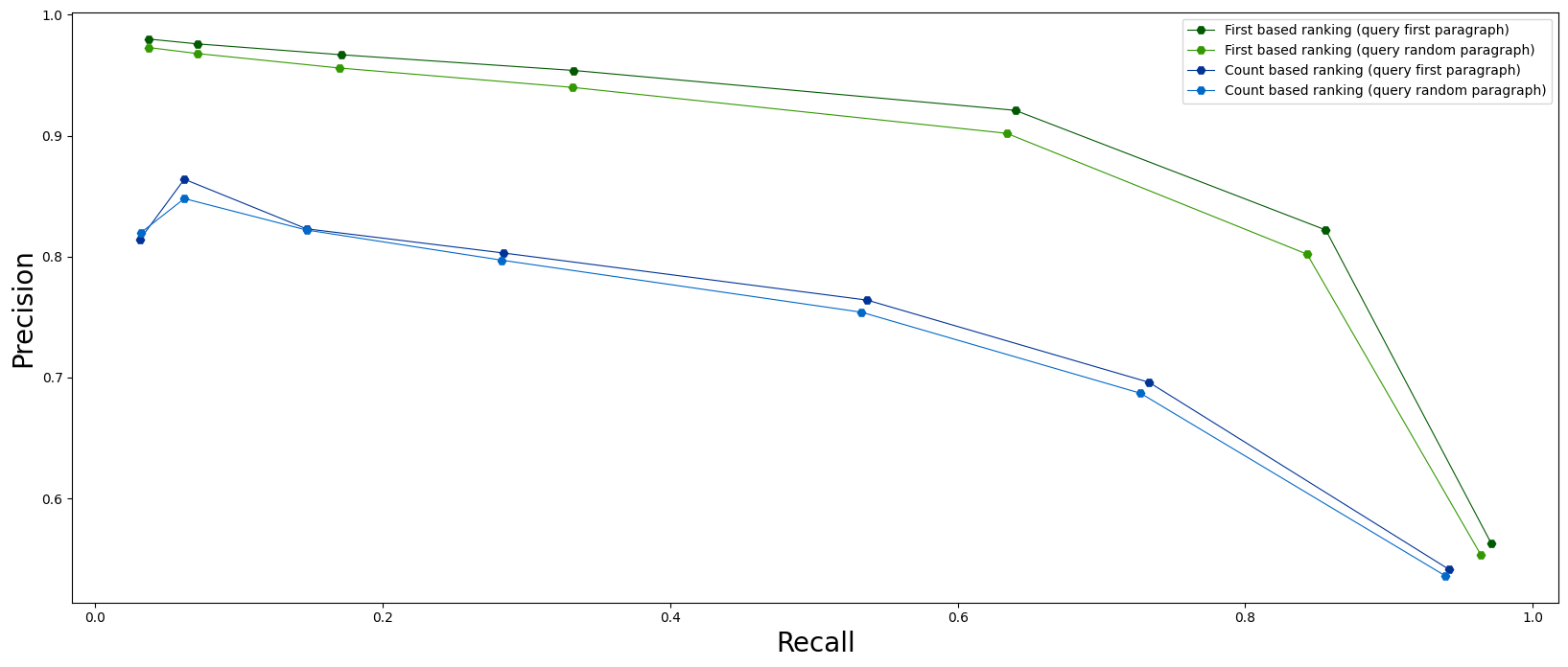}
    \caption{Different configurations for DVS on the test set}
    \label{fig:bert}
\end{figure}

For our test set, the recall-precision graphs are shown in Figure \ref{fig:tfidf} and Figure \ref{fig:bert}. In both Figures, it's apparent that deriving a document ranking from its \textit{highest} ranked paragraph outperforms ranking documents based on \textit{counting} the paragraphs. Moreover, for both methods querying the first paragraph of a document slightly outperforms querying random paragraphs. Next, for the TF-IDF-based method specifically, the experiments on the document level outperform the experiments on the paragraph level.

\begin{figure}[p]
    \centering
    \includegraphics[width=\textwidth]{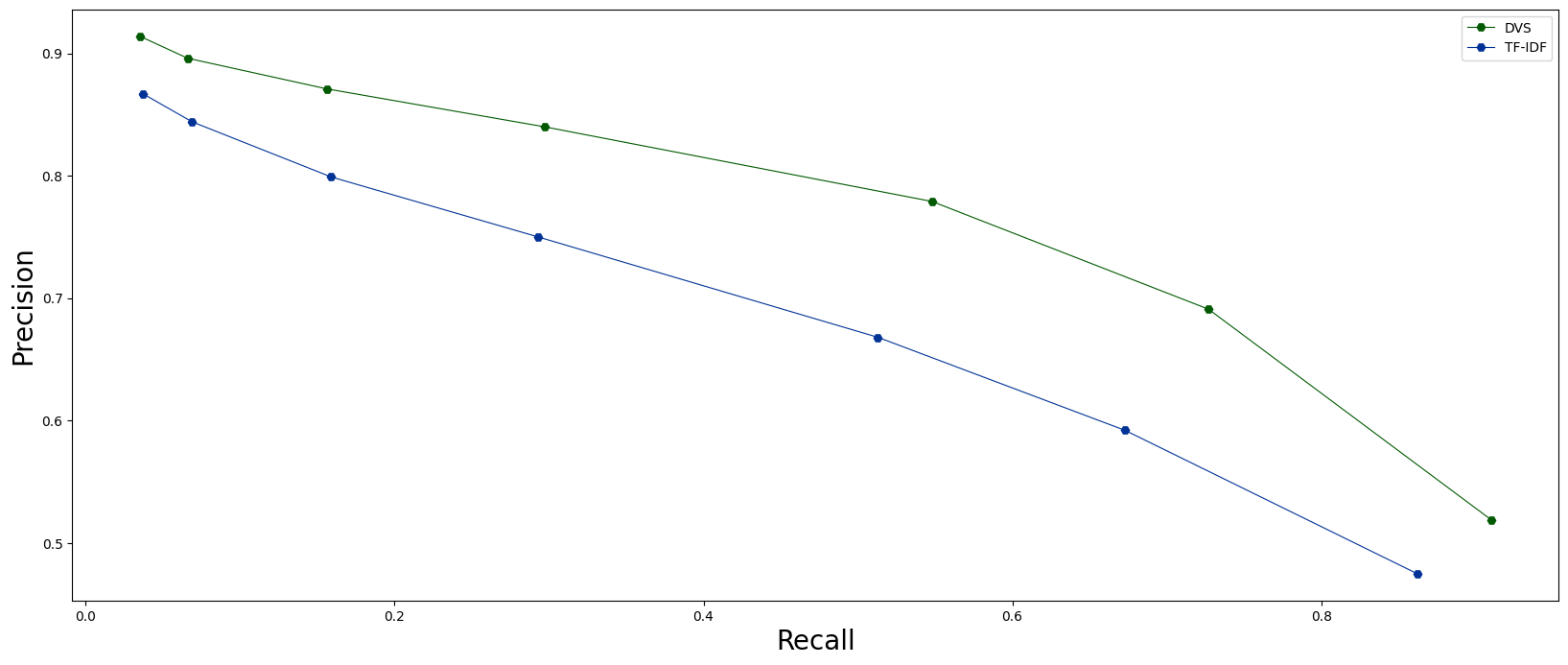}
    \caption{Identified configurations of DVS and TF-IDF-based approach on the ambiguous set}
    \label{fig:ambi}
\end{figure}

Finally, when comparing the performance of the optimal configuration of both approaches, DVS (with a first-based ranking) outperforms all TF-IDF-based approaches. Next, when comparing the optimal configurations of the approaches on the \textit{ambiguous} set, we again see that DVS outperforms the TF-IDF-based approach (see Figure \ref{fig:ambi}).  

\subsection{Relevance feedback}
For the second set of experiments, we experimented with different feedback strategies using the identified text similarity method (DVS). For these experiments, the average number of iterations (and standard deviation) needed to achieve 80\% recall for all methods and datasets is shown in Table \ref{tab:feedback-80}. Note, every iteration translates to a review effort of 10 paragraphs. Similar to the previous experiments, the results are available on the test set and the ambiguous set.

\begin{table}[tb]
    \centering
    \resizebox{\textwidth}{!}{%
    \begin{tabular}{|l|l|l|l|l|}
    \hline
    \textbf{Feedback strategy} & \textbf{\begin{tabular}[c]{@{}l@{}}Mean \\ (Test)\end{tabular}} & \textbf{\begin{tabular}[c]{@{}l@{}}Std Dev \\ (Test)\end{tabular}} & \textbf{\begin{tabular}[c]{@{}l@{}}Mean \\ (Ambiguous)\end{tabular}} & \textbf{\begin{tabular}[c]{@{}l@{}}Std Dev \\ (Ambiguous)\end{tabular}} \\ \hline \hline
    \textbf{No feedback}          & 33.36                                                          & 8.84                                                              & 46.77                                                               & 10.04                                                                  \\ \hline
    \textbf{Keyword expansion} & 32.24                                                          & 8.03                                                              & 44.23                                                               & 9.94                                                                   \\ \hline 
    \textbf{Rocchio ($\alpha=0.5$, $\beta=0.5$)} &  29.50  &  3.79  &  37.98  &   5.11     \\ \hline \hline 
    \textbf{Average}           & 29.43                                                          & 1.51                                                              & 32.65                                                               & 2.97                                                                   \\ \hline
    \textbf{Average (amp)}     & 30.17                                                          & 1.03                                                              & 32.99                                                               & 1.80                                                                   \\ \hline
    \textbf{Sum}               & \textit{\textbf{28.29}}                                                          & 0.75                                                              & \textit{\textbf{29.41}}                                                               & 2.13                                                                  \\ \hline
    \textbf{Sum (amp)}         & 28.46                                                          & \textit{\textbf{0.50}}                                                              & 30.54                                                               & \textit{\textbf{1.31}}                                                                   \\ \hline \hline
    \textbf{Average}           & 31.54                                                          & 1.14                                                              & 38.68                                                               & 4.41                                                                   \\ \hline
    \textbf{Average (amp)}     & 32.81                                                          & 1.08                                                              & 38.41                                                               & 2.64                                                        \\ \hline
    \textbf{Sum}               & 32.80                                                          & 1.12                                                              & 38.80                                                               & 5.28                                                                  \\ \hline
    \textbf{Sum (amp)}         & 32.20                                                          & 0.87                                                             & 38.20                                                               & 3.77                                                                   \\ \hline
    \end{tabular}
    }
    \caption{Iterations needed to achieve 80\% recall (baselines are in the top cells, cumulative approaches are in the middle cells, non-cumulative approaches are in the bottom cells).}
    \label{tab:feedback-80}
\end{table}

First, the results on the test set. Here, it's apparent that the feedback methods based on vector operations require fewer iterations and have a lower standard deviation that the baseline methods. The best performing feedback method is summing the vectors. Next, the results on the ambiguous set. Here, all similarity methods require more iterations to reach 80\% recall and have a higher standard deviation than they have on the test set. Still, feedback methods based on summing the vectors (cumulatively) gives the best results. 

It's apparent that cumulative feedback methods outperform non-cumulative feedback methods for all vector-based relevance strategies. Moreover, the difference between averaging and summing vectors is smaller when using non-cumulative strategies.   

A noteworthy finding when comparing the results from the test set and the ambiguous set is the gap between the minimal baseline method (of no feedback) and the optimal feedback method (of cumulatively summing the vectors). On the test set, the reduction of review effort (measured in the number of iterations to achieve 80\% recall) is 17.85\%. On the ambiguous set however, this reduction of review effort equals 59.04\%. Another noteworthy finding is that amplifying the feedback to sibling paragraphs reduces the standard deviation, but not the number of iterations needed to achieve 80\% recall.

Finally, for all methods we measured average time taken per iteration. The results for this are shown in Table \ref{tab:execution-time-relevance-feedback}. These results show that relevance feedback strategies based on vector operations (average and sum) add little latency to the experiment compared to no feedback. Still, amplifying feedback to sibling paragraphs does add some latency to the experiment for both averaging and summing vectors. However, keyword expansion adds the most latency to the experiments. 

\begin{table}[t]
\centering
\begin{tabular}{|l|l|}
\hline
\textbf{Strategy}          & \textbf{Average time per iteration (in seconds)} \\ \hline \hline
\textbf{No feedback}       & 0.02298                                          \\ \hline
\textbf{Average}           & 0.02386                                          \\ \hline
\textbf{Sum}               & 0.02417                                          \\ \hline
\textbf{Rocchio}           & 0.02901                                          \\ \hline
\textbf{Sum (amp)}         & 0.04982                                          \\ \hline
\textbf{Average (amp)}     & 0.05097                                          \\ \hline
\textbf{Keyword expansion} & 0.12093                                          \\ \hline
\end{tabular}
\caption{Average execution times for different relevant feedback strategies}
\label{tab:execution-time-relevance-feedback}
\end{table}

\section{Discussion}
\label{sec:discussion}

This section is a discussion of the results. First, we interpret the results for our two main experiments. Second, we discuss the limitations of our research and the resulting suggestions for future work.

\subsection{Text similarity methods}
For all of text similarity methods, some common denominators emerge from the results. First, querying the first paragraph gives better results that querying a random paragraph. This coincides with findings from related work, since prior research found that the effectiveness of ``query by document" approaches depends on the prevalence of relevant terms in the queried text \cite{Eugene}. Combining this finding with our results, it's probable that the first paragraphs in the RCV-1 v2 dataset outperform random paragraphs because they have a higher prevalence of relevant terms. This property could be exclusive to the news articles used in our research, and therefore might not apply to the data used in other domains 

Another commonality between the methods is related to the paragraph-based document rankings. Here, we see that for all methods ranking documents based on the first paragraph in the ranking gives better results than querying documents based on counting their paragraphs in the ranking. Potentially, this could be related to differences in the number of paragraphs per document. Because, if a document only has one paragraph, then it's always bound to be at the bottom of a count-based ranking. Regardless of how related/similar that document is. 

As for comparing the different text similarity methods, both sets of experiments show that DVS outperforms the TF-IDF-based approach. Considering our DVS approach is based on BERT instead of TF-IDF, this finding makes sense. Because, BERT's bidirectional self-attention mechanism \cite{Devlin} enables it to understand more ambiguous and context-sensitive language properties. For example, the usage of synonymy, homonymy and referrals. 

\subsection{Relevance feedback}
For the second experiment, the best performing text similarity method (DVS, with a first-based document ranking) was implemented for relevance feedback. For both the test set and the ambiguous set, the experiments show that relevance feedback methods based on cumulatively summing the vectors reduce review effort the most. Interestingly, this improvement does not seem to come at a computational cost. Since Table \ref{tab:execution-time-relevance-feedback} shows that the execution time of these methods (without feedback amplification to sibling paragraphs) is fairly similar to our minimal baseline method (of no feedback). Note, when optimizing for a low standard deviation, amplifying feedback to sibling paragraphs is beneficial, which does add latency to the experiment. 

\subsection{Relevance}
First, with regards to implementation, our results show that review effort can be decreased using text similarity-based relevance feedback methods. An important side note in this finding is that relevance feedback accomplishes this through only re-ranking documents. As a result, this strategy does not produce false negatives automatically without the awareness of the user (in contrast to an active learning based approach). This is particularly important in use-cases that require a high recall. 

Second, with regards to scientific novelty and contributions, we should state that the concept of relevance feedback has been studied before \cite{Zhang}. However, certain parts within our implementation are (to our knowledge) novel and therefore contribute to science. 

First, the evaluation of different paragraph-based document rankings contribute to the domain of paragraph-based document-to-document retrieval. Given the rise of large language models (that are generally limited to the paragraph level \cite{Liu-2}) and the fact that relevant information can be exclusive to a specific part of a document \cite{Adebayo}, these findings are applicable beyond the technologies/embeddings used in this research. 

Second, our results show that the usage of sibling paragraphs in relevance feedback can reduce the standard deviation of review effort. To our knowledge, this technique and finding is novel. Moreover, a more ``stable" reduction of review effort could be favorable in real-world scenarios. Hence, this finding is not just novel, but also applicable.

\subsection{Limitations}
The size of the dataset (only 300 articles per topic) is smaller than most real-world information retrieval scenarios. Moreover, our data only consists of news articles. Therefore, certain findings in our research (e.g., the fact that querying the first paragraph slightly outperforms querying random paragraphs) might not apply on other datasets. Still, given the fact that the RCV-1 v2 dataset is commonly used as a benchmark dataset in information retrieval \cite{Deolalikar}, the results are still an adequate indication of our method's performance.  

\subsection{Future work}
Our first suggestion for future work is related to the size of our dataset. As mentioned, due to computational constraints we only sampled 300 articles per topic. However, datasets in real-world information retrieval applications can be much larger than that. As a result, future work could run these methods on larger datasets to research their scalability.

Next, this research uses Solr's ``MoreLikeThis" functionality to increase the speed of our TF-IDF-based similarity experiments. Meaning, we didn't use any vector space scoring to compute the similarities between the TF-IDF vectors. The reasoning behind this is that computing the (cosine) similarities between \textit{n} TF-IDF vectors results in quadratic (O(n$^2$)) time and space complexity, which is simply not viable in a real-world application. 

However, recent innovations have made it possible to compute the pairwise similarities between (sparse) vectors much faster. An example of this is the ChunkDot Python library \cite{ChunckDot}, which splits the TF-IDF matrix into chunks and computes the similarities in parallel. Future work could use this innovation to experiment with TF-IDF-based cosine similarity as an additional text similarity method. 

\section{Conclusion}
\label{sec:conclusion}

This research aimed to evaluate the impact of changing the (text similarity-based) relevance rankings based on relevance feedback. For this, the first research question was formulated as follows: what text similarity methods are most suitable for relevance feedback? Our results show that the most suitable text similarity method for this is DVS (using BERT-based dense-vector representations) where the highest ranked paragraph determines the document ranking. 

Next, the second research question was formulated as follows: to what extent can relevance feedback help to reduce review effort? Here, our results show that (compared to processing no relevance feedback) the relevance feedback method identified in this research reduces review effort between 17.85\% and 59.04\%, given a target recall level of 80\%. 

Given the recall-oriented nature of many information retrieval applications \cite{Song}, the results for the relevance feedback experiments are very encouraging. Since, in contrast to an active learning based strategy (which typically used for this purpose), this approach reduces review effort through only re-ranking documents. As a result, there are no false negatives created without the awareness of the user. 

\bibliographystyle{splncs04}
\bibliography{bibliography}

\begin{thebibliography}{10}
\providecommand{\url}[1]{\texttt{#1}}
\providecommand{\urlprefix}{URL }
\providecommand{\doi}[1]{https://doi.org/#1}

\bibitem{Adebayo}
Adebayo, K.J.: Multimodal Legal Information Retrieval. Ph.D. thesis, University
  of Luxembourg, Luxembourg (2018)

\bibitem{ChunckDot}
Agundez, R.: {ChunckDot Python library} (May 2023),
  \url{https://pypi.org/project/chunkdot/}

\bibitem{RCV-2}
Amini, Massih-Reza \&~Goutte, C.: {Reuters RCV1 RCV2 Multilingual, Multiview
  Text Categorization Test collection}. UCI Machine Learning Repository (2013)

\bibitem{Arampatzis}
Arampatzis, A., Peikos, G., Symeonidis, S.: Pseudo relevance feedback
  optimization. Information Retrieval Journal pp. 269--297 (2021)

\bibitem{Cai}
Cai, T., He, Z., Hong, C., Zhang, Y., Ho, Y.L., Honerlaw, J., Geva, A.,
  Panickan, V.A., King, A., Gagnon, D.R., et~al.: Scalable relevance ranking
  algorithm via semantic similarity assessment improves efficiency of medical
  chart review. Journal of Biomedical Informatics p. 104109 (2022)

\bibitem{Chai}
Chai, C.P.: Comparison of text preprocessing methods. Natural Language
  Engineering pp. 509--553 (2023)

\bibitem{Chandrasekaran}
Chandrasekaran, D., Mago, V.: Evolution of semantic similarity{\textemdash}a
  survey. {ACM} Computing Surveys pp. 1--37 (mar 2022)

\bibitem{Deolalikar}
Deolalikar, V.: {How valuable is your data? A quantitative approach using data
  mining}. In: 2015 IEEE International Conference on Big Data (Big Data). pp.
  1248--1253. IEEE (2015)

\bibitem{Devlin}
Devlin, J., Chang, M., Lee, K., Toutanova, K.: {BERT:} pre-training of deep
  bidirectional transformers for language understanding. CoRR
  \textbf{abs/1810.04805} (2018)

\bibitem{Lucene}
Foundation, A.S.: Apache {Lucene} - scoring (2011),
  \url{http://lucene.apache.org/java/340/scoring.html}

\bibitem{Gorman}
Gorman, J., Curran, J.R.: Scaling distributional similarity to large corpora.
  In: Proceedings of the 21st International Conference on Computational
  Linguistics and 44th Annual Meeting of the Association for Computational
  Linguistics. pp. 361--368 (2006)

\bibitem{Grossman}
Grossman, M.R., Cormack, G.: {Continuous active learning for TAR}. The Journal
  pp.~1--7 (2016)

\bibitem{Jaccard}
Jaccard, P.: Nouvelles recherches sur la distribution florale. Bull. Soc. Vaud.
  Sci. Nat. pp. 223--270 (1908)

\bibitem{Hampp}
Joshi, S., Contractor, D., Ng, K., Deshpande, P.M., Hampp, T.: {Auto-Grouping
  Emails for Faster e-Discovery}. Proc. VLDB Endow. p. 1284–1294 (June 2020)

\bibitem{Li}
Li, H., Mourad, A., Zhuang, S., Koopman, B., Zuccon, G.: Pseudo relevance
  feedback with deep language models and dense retrievers: Successes and
  pitfalls. ACM Transactions on Information Systems pp. 1--40 (2023)

\bibitem{Liu-2}
Liu, Y., Han, T., Ma, S., Zhang, J., Yang, Y., Tian, J., He, H., Li, A., He,
  M., Liu, Z., et~al.: {Summary of ChatGPT research and perspective towards the
  future of large language models}. arXiv preprint arXiv:2304.01852  (2023)

\bibitem{Malkov}
Malkov, Y.A., Yashunin, D.A.: Efficient and robust approximate nearest neighbor
  search using hierarchical navigable small world graphs. IEEE Computer Society
  p. 824–836 (apr 2020)

\bibitem{Manning}
Manning, C.D.: Introduction to information retrieval. Syngress Publishing
  (2008)

\bibitem{Piramuthu}
Piramuthu, O.B.: Multiple choice online algorithms for technology-assisted
  reviews. In: Proceedings of the 38th ACM/SIGAPP Symposium on Applied
  Computing. pp. 639--645 (2023)

\bibitem{Rahutomo}
Rahutomo, F., Kitasuka, T., Aritsugi, M.: Semantic cosine similarity. In: The
  7th international student conference on advanced science and technology
  ICAST. p.~1 (2012)

\bibitem{Reimers}
Reimers, N., Gurevych, I.: {Sentence-BERT: Sentence Embeddings using Siamese
  BERT-Networks}. CoRR pp. 3980--3990 (2019)

\bibitem{Rocchio}
Rocchio, J.J.: {Document Retrieval System-Optimization and Evaluation}. DIR
  2009 Dutch-Belgian Information Retrieval Workshop p.~99 (2009)

\bibitem{Hedin}
Roitblat, H.L.: {Probably Reasonable Search in eDiscovery}. arXiv e-prints
  p.~2201 (2022)

\bibitem{Sanchez}
S{\'a}nchez, D., Batet, M.: A semantic similarity method based on information
  content exploiting multiple ontologies. Expert Systems with Applications pp.
  1393--1399 (2013)

\bibitem{Song}
Song, J.J., Lee, W., Afshar, J.: An effective high recall retrieval method.
  Data \& Knowledge Engineering  \textbf{123},  101603 (2019)

\bibitem{Vaswani}
Vaswani, A., Shazeer, N., Parmar, N., Uszkoreit, J., Jones, L., Gomez, A.N.,
  Kaiser, {\L}., Polosukhin, I.: Attention is all you need. Advances in neural
  information processing systems  (2017)

\bibitem{Eugene}
Yang, E., Lewis, D.D., Frieder, O., Grossman, D.A., Yurchak, R.: {Retrieval and
  Richness when Querying by Document.} In: DESIRES. pp. 68--75 (2018)

\bibitem{Zhang}
Zhang, H., Cormack, G.V., Grossman, M.R., Smucker, M.D.: Evaluating
  sentence-level relevance feedback for high-recall information retrieval.
  Information Retrieval Journal pp. 1--26 (2020)

\end{thebibliography}
\end{document}